\documentclass{article}

\usepackage[tight]{subfigure}
\usepackage{graphicx}

\usepackage{appendix}

\usepackage[authoryear]{natbib}

\usepackage[algo2e,ruled,vlined]{algorithm2e}
\usepackage{hyperref}

\usepackage[accepted]{icml2013}

\usepackage{amsthm}
\usepackage{amsfonts}
\usepackage{amsmath}
\usepackage{amssymb}
\usepackage{wasysym}
\usepackage{mdwlist}
\usepackage{xspace}
\usepackage{enumitem}

\newcommand{\ben}{\begin{enumerate*}}
\newcommand{\een}{\end{enumerate*}}
\newcommand{\beq}{\begin{eqnarray}}
\newcommand{\eeq}{\end{eqnarray}}
\newcommand{\bit}{\begin{itemize*}}
\newcommand{\eit}{\end{itemize*}}
\newcommand{\nn}{\nonumber}

\newcommand{\reminder}[1]{{\color{red} *** #1 *** }}

\newcommand{\hide}[1]{}

\newtheorem{defn}{Definition} 
\newtheorem{proposition}{Proposition} 
\newtheorem{theorem}{Theorem} 

\newcommand{\myvec}[1]{{#1}}

\newcommand{\mymat}[1]{{\mathbf #1}}

\newcommand{\indep}[1]{\perp}

\DeclareMathOperator{\Cauchy}{Cauchy}
\DeclareMathOperator{\obs}{obs}
\DeclareMathOperator{\std}{std}
\DeclareMathOperator{\trace}{trace}
\DeclareMathOperator{\update}{update}

\newcommand{\appropto}{\mathrel{\vcenter{
  \offinterlineskip\halign{\hfil$##$\cr
    \propto\cr\noalign{\kern2pt}\sim\cr\noalign{\kern-2pt}}}}}

\newcommand{\N}{\mathcal{N}}

\newcommand{\secref}[1]{Section~\ref{#1}}

\newcommand{\algref}[1]{Algorithm~\ref{#1}}
\newcommand{\figref}[1]{Figure~\ref{#1}}

\renewcommand{\eqref}[1]{Equation~(\ref{#1})}

\newcommand{\eq}{{\,{=}\,}}     
\newcommand{\given}{\!\mid\!}

\renewcommand{\reminder}[1]{} 

\icmltitlerunning{The Extended Parameter Filter}

\begin{document}

\twocolumn[
\icmltitle{The Extended Parameter Filter}

\icmlauthor{Yusuf B. Erol$^\dagger$}{yberol@eecs.berkeley.edu}
\icmlauthor{Lei Li$^\dagger$}{leili@cs.berkeley.edu}
\icmlauthor{Bharath Ramsundar}{rbharath@stanford.edu}
\icmladdress{Computer Science Department, Stanford University}
\icmlauthor{Stuart Russell$^\dagger$}{russell@cs.berkeley.edu}
\icmladdress{$^\dagger$EECS Department, University of California, Berkeley}

\icmlkeywords{sequential importance sampling, particle filtering, taylor approximation, static parameter estimation, Taylor approximation}
\vskip 0.3in
]

\begin{abstract}
\label{sec:abstract}
The parameters of temporal models, such as dynamic Bayesian networks,
may be modelled in a Bayesian context as static or atemporal variables
that influence transition probabilities at every time step.
Particle filters fail for models that include such variables, while
methods that use Gibbs sampling of parameter variables may incur a
per-sample cost that grows linearly with the length of the observation
sequence. \citet{storvik2002particle} devised a method for incremental computation
of exact sufficient statistics that, for some cases, reduces
the per-sample cost to a constant.  In this paper, we demonstrate
a connection between Storvik's filter and a Kalman filter in parameter space
and establish more general conditions under which Storvik's filter works.
Drawing on an analogy to the extended Kalman filter, we develop and
analyze, both theoretically and experimentally, a Taylor approximation
to the parameter posterior that allows Storvik's method to be applied
to a broader class of models.
Our experiments on both synthetic examples and real applications show improvement over existing methods.

\end{abstract}

\section{Introduction}
\label{sec:intro}

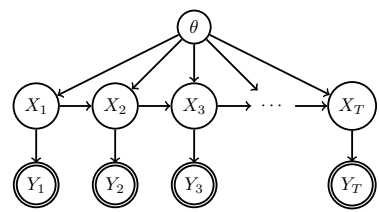
\begin{figure}[tb]
\centering
\hide{
  \begin{tikzpicture}
      [edge from parent/.style={draw,->}, 
             line width=1.0pt,
             ]
      \tikzstyle{every node}=[circle,draw]
      \node {$\theta$} 
        child { node(x1) {$X_1$}
                 child { node[double] {$Y_1$}  }
               }
        child { node(x2) {$X_2$}
                 child { node[double] {$Y_2$}  }
               }
        child { node(x3) {$X_3$}
                 child { node[double] {$Y_3$}  }
               }
        child { node(dots) [draw=white] {$\cdots$}}
        child { node(xt) {$X_T$}
                 child { node[double] {$Y_T$}  }
               }
      ;
       \path[->] {
            (x1) edge (x2)
            (x2) edge (x3)
            (x3) edge (dots) 
            (dots)  edge (xt) 
          }
          ;
  \end{tikzpicture}
}
 \includegraphics[scale=0.7]{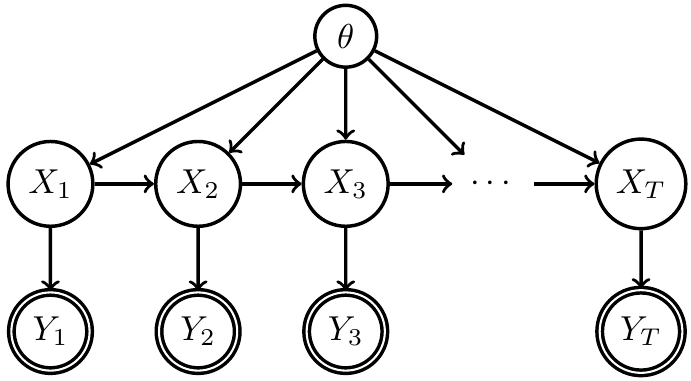}
 \vspace{-0.1in}
 \caption{A state-space model with static parameters $\theta$. $X_{1:T}$ are latent states and $Y_{1:T}$ are observations.}
\label{fig:graphmodel}
\end{figure}

Dynamic Bayesian networks are widely used to model 
the processes underlying sequential data such as
speech signals, financial time series, 
genetic sequences, and medical or physiological signals. 
State estimation or filtering---computing the posterior 
 distribution over the state of a partially observable Markov
process from a sequence of observations---is one of the most widely studied
problems in control theory, statistics and AI. Exact filtering
is intractable except for certain special cases (linear--Gaussian
models and discrete HMMs), but approximate filtering using
the {\em particle filter} (a sequential Monte Carlo
method) is feasible in many real-world applications \citep{arulampalam2002tutorial,doucet2011tutorial}. 
In the machine learning context, model parameters may be represented by
static parameter variables that define the transition and sensor model probabilities of the Markov process, but do not themselves change over time (\figref{fig:graphmodel}).
The posterior parameter distribution (usually) converges to a delta function at the true value
in the limit of infinitely many observations.
Unfortunately, particle filters fail for such models: the algorithm
samples parameter values for each particle at time $t\eq 0$, but these remain fixed;
over time, the particle resampling process removes all but one set of values; and these
are highly unlikely to be correct. The degeneracy problem is especially severe in high-dimensional
parameter spaces, whether discrete or continuous.
Hence, although learning requires inference, the most successful inference algorithm for temporal models is inapplicable.

\citet{kantas2009overview} and \citet{carvalho2010particle} describe several algorithms that have been proposed to solve this degeneracy problem,
but the issue remains open because known algorithms either suffer from
bias or computational inefficiency. For example, the ``artificial dynamics'' approach
\citep{liu2001combined} introduces a stochastic transition model for the parameter variables, allowing exploration of the parameter space, but this may result in biased estimates.
Online EM algorithms \citep{andrieu2005line} provide only point
estimates of static parameters, may converge to local optima, and are
biased unless used with the full smoothing distribution.  The particle
MCMC algorithm \citep{andrieu2010particle} converges to the true posterior, but
requires computation growing with $T$, the length of the data
sequence.

The resample-move algorithm \citep{gilks2001following} includes Gibbs sampling of
parameter variables---that is, in \figref{fig:graphmodel}, $P(\theta \given X_1,\ldots,X_T)$.
This method requires $O(T)$ computation per sample, leading \citeauthor{gilks2001following}
to propose a sampling rate proportional to $1/T$ to preserve constant-time updates.
\citet{storvik2002particle} and \citet{polson2008practical} observe that 
a fixed-dimensional sufficient statistic (if one exists) for $\theta$ can be updated in constant time.
Storvik describes an algorithm for a specific family of linear-in-parameters transition models.
 
We show that Storvik's algorithm is a special case of the Kalman
filter in parameter space and identify a more general class of {\em
separable} systems to which the same approach can be applied.  By
analogy with the extended Kalman filter, we propose a new algorithm,
the {\em extended parameter filter} (EPF), that computes a separable
approximation to the parameter posterior and allows a
fixed-dimensional (approximate) sufficient statistic to be maintained.
The method is quite general: for example, with a polynomial
approximation scheme such as Taylor expansion any analytic posterior
can be handled.


\secref{sec:storvik} briefly reviews particle filters and Storvik's method and
introduces our notion of separable models. \secref{sec:algorithm}
describes the EPF algorithm, and \secref{sec:poly_general}
discusses the details of a polynomial approximation scheme for arbitrary
densities, which \secref{sec:online_gibbs_approx} then applies to estimate posterior distributions of
static parameters. \secref{sec:experiment} provides empirical results
comparing the EPF to other algorithms. All details of proofs are given 
in the appendix of the full version \cite{erol2013extended}.

\section{Background}
\label{sec:storvik}
In this section, we review state-space dynamical models and 
the basic framework of approximate filtering algorithms. 

\subsection{State-space model and filtering}
Let $\Theta$ be a parameter space for a partially observable
Markov process $\left\{X_{t} \right\}_{t \geq 0}, \left \{Y_t \right \}_{t \geq
0}$ as shown in \figref{fig:graphmodel} and defined as follows:
\begin{align}
	X_0 & \sim p(x_0\given \theta) \\
    X_t \given x_{t-1} &\sim p(x_t \given x_{t-1}, \theta) \\
	Y_t \given x_t &\sim p(y_t \given x_t, \theta)
	\label{eq:state-space-model}
\end{align} 
Here the state variables $X_t$ are unobserved and the observations $Y_{t}$ are assumed conditionally independent of
other observations given $X_{t}$.  We assume in this section that states $X_t$, observations $Y_t$, and parameters $\theta$ are
real-valued vectors in $d$, $m$, and $p$ dimensions respectively. 
Here both the transition  and sensor models are 
parameterized by $\theta$. 
For simplicity, we will assume in the following sections that only the transition model
is parameterized by $\theta$; however, the results in this paper can
be generalized to cover sensor model parameters.

The filtering density $p(x_t \given y_{0:t}, \theta)$ obeys the following
recursion:
\begin{align}
\label{eq:bayes1}
&	p(x_t  \given y_{0:t}, \theta) =\frac{p(y_t \given x_t, \theta) p(x_t \given
    y_{0:t-1}, \theta) }{ p(y_t  \given y_{0:t-1}, \theta) } \notag \\
  &=  \frac{p(y_t \given x_t, \theta)  }{ p(y_t  \given y_{0:t-1}, \theta) } \int p(x_{t-1} \given
        y_{0:t-1}, \theta) p(x_{t}\given x_{t-1}, \theta) d x_{t-1}
\end{align}
where the update steps for $p(x_t  \given y_{0:t-1}, \theta)$ and $p(y_t \given
y_{0:t-1}, \theta)$ involve the evaluation of integrals that are not in general
tractable.


\subsection{Particle filtering}
With known parameters, particle filters can approximate the posterior distribution over the hidden state $X_{t}$
by a set of samples. The canonical example is the sequential importance sampling-resampling  algorithm (SIR)  (\algref{alg:SIR}). 

\begin{algorithm2e}[tb]
\KwIn{$N$: number of particles;\\
$y_0,\dots, y_T$: observation sequence}
\KwOut{$\bar{x}_{1: T}^{1: N}$}
initialize $\left\{x_0^i\right\}$ \;
\For{$t=1,\ldots,T$} {
	\For{$i=1,\dots,N$}{
	        sample $x_{t}^i \sim p(x_t \given x_{t-1}^i)$\;
	        $w_t^i \leftarrow p(y_t \given x_t^i)$\;
	}
	sample $\left\{\frac{1}{N},\bar{{x}}_t^i\right\}\leftarrow  $Multinomial$\left\{w_t^i,{x}_t^i\right\}$\;
	$\left\{{x}_t^i\right\}\leftarrow \left\{\bar{{x}}_t^i\right\}$\;
}
\caption{Sequential importance sampling-resampling  (SIR)}
\label{alg:SIR}
\end{algorithm2e}

The SIR filter has various appealing properties. It is modular, efficient, and easy to implement. The  filter takes constant time per update, regardless of time $T$, and as the number of particles $N
\to \infty$, the empirical filtering density converges to the true
marginal posterior density under suitable assumptions.

Particle filters can accommodate unknown parameters by adding parameter variables into the state
vector with an ``identity function'' transition model. As noted in \secref{sec:intro} this approach leads to degeneracy problems---especially
for high-dimensional parameter spaces. To ensure that {\em some} particle has
initial parameter values with bounded error, the number of particles must grow
exponentially with the dimension of the parameter space.

\subsection{Storvik's algorithm}
To avoid the degeneracy problem, \citet{storvik2002particle} modifies the SIR algorithm by
 adding a Gibbs sampling step for $\theta$ conditioned 
 on the state trajectory in each particle (see \algref{alg:Storvik}). The algorithm is developed in the SIS framework
 and consequently inherits the theoretical guarantees of SIS. \reminder{the LIP
 stuff below gets repeated} Storvik considers unknown parameters in the
 state evolution model and assumes a perfectly known sensor model. His analysis
 can be generalized to unknown sensor models.

\hide{For models with transition equation linear to the parameters with Gaussian noises 
 (but can be nonlinear to the states), the conditional
 distribution of $p(\theta \given x_{0:t-1}, y_{0:t-1})$
can be calculated within time complexity irrelevant to $t$. }
Storvik's approach becomes efficient in an on-line setting when a
fixed-dimensional sufficient statistic ${S}_t$ exists for the static parameter 
(\textit{i.e.}, when $p(\theta | {x}_{0:t})=p(\theta | {S}_t)$ holds). 
The important property of this algorithm is that the parameter value simulated
at time $t$ does not depend on the values simulated previously. This property
prevents the impoverishment of the parameter values in particles. 

\begin{algorithm2e}[tb]
\KwIn{$N$: number of particles;\\
$y_0,\dots, y_T$: observation sequence}
\KwOut{$\bar{x}_{1:T}^{1:N}$, $\theta^{1:N}$}
initialize $\left\{x_0^i\right\}$\;
\For{$t=1,\ldots,T$} {
	\For{$i=1,\dots,N$}{
	        sample $\theta^i \sim p(\theta | {x}_{0:t-1}^i)$\;
	        sample ${x}_{t}^i \sim p({x_t}|{x}_{t-1}^i,\theta^i)$\;
	        $w^i \leftarrow p({y}_t | {x}_t^i)$\;
	}
	sample $\left\{\frac{1}{N},\bar{{x}}_t^i\right\}\leftarrow  $Multinomial$\left\{w_t^i,{x}_t^i\right\}$\;
	$\left\{{x}_t^i\right\}\leftarrow \left\{\bar{{x}}_t^i\right\}$\;
}
\caption{Storvik's filter.}
\label{alg:Storvik}
\end{algorithm2e}


One limitation of the algorithm is that it can only be 
applied to models with fixed-dimensional sufficient statistics.
However, \citet{storvik2002particle} analyze the 
sufficient statistics for a specific family.

\citet{storvik2002particle} shows how to obtain a sufficient statistic in the context of 
what he calls the {\em Gaussian system process}, a transition model satisfying the equation
\beq
	\myvec{x}_t=\mymat{F}_t^T\theta+\epsilon_t, \>\> \epsilon_t \sim N(0,\mymat{Q})
\label{eq:storvik_model}
\eeq
where $\theta$ is the vector of unknown parameters with a prior of
$N(\theta_0,\mymat{C}_0)$ and
$\mymat{F}_t=\mymat{F}({x}_{t-1})$ is a matrix where elements are
possibly nonlinear functions of ${x}_{t-1}$. An arbitrary but
known observation model is assumed. Then the standard theory states
that $\theta \given {x}_{0:t} \sim
N(\myvec{m}_t,\mymat{C}_t)$ where the recursions for the mean and
the covariance matrix are as follows:
\beq
\label{eq:storvik_updates}
	\mymat{D}_t&=&\mymat{F}_t^T \mymat{C}_{t-1} \mymat{F}_t+\mymat{Q} \nn \\
	\mymat{C}_t&=&\mymat{C}_{t-1}-\mymat{C}_{t-1}\mymat{F}_t\mymat{D}_t^{-1}\mymat{F}_t^T\mymat{C}_{t-1} \nn \\
	\myvec{m}_t&=&\myvec{m}_{t-1}+\mymat{C}_{t-1}\mymat{F}_t\mymat{D}_t^{-1}(\myvec{x}_t-\mymat{F}_t^T\myvec{m}_{t-1}) 
\eeq
Thus, $\myvec{m}_t$ and $\mymat{C}_t$ constitute a fixed-dimensional sufficient statistic for $\theta$.

These updates are in fact a special case of Kalman filtering applied to the parameter space.
Matching terms with the standard KF update equations~\citep{kalman1960new}, we find that
the transition matrix for the KF is the identity matrix,
the transition noise covariance matrix is the zero matrix,
the observation matrix for the KF is $\mymat{F}_t$,
and the observation noise covariance matrix is $\mymat{Q}$.
This correspondence is of course what one would expect, since the true parameter values are fixed (i.e., an identity transition). 
See the supplementary material \cite{erol2013extended} for the derivation.

\subsection{Separability}

In this section, we define a condition under which there
exist efficient updates to parameters. 
Again, we focus on the state-space model as described in
\figref{fig:graphmodel} and \eqref{eq:state-space-model}. The model in \eqref{eq:state-space-model} can also be expressed as
\beq
	x_t = f_{\theta}(x_{t-1}) + v_t \nn \\
	y_t = g(x_t) + w_t
\eeq
for some suitable $f_{\theta}$, $g$, $v_t$, and $w_t$.

\begin{defn}
A system is \emph{separable} if the transition function $f_{\theta}(x_{t-1})$
can be written as $f_{\theta}(x_{t-1})=l(x_{t-1})^Th(\theta)$ for some
$l(\cdot)$ and $h(\cdot)$ and if the stochastic i.i.d. noise $v_t$ has log-polynomial density.
\end{defn}

\hide{
\begin{defn}
A transition model $p(x_t \given x_{t-1}, \theta)$ is \em{separable} if it can be written as $p(x_t \given x_{t-1}, \theta) =  g(x_t, x_{t-1})^T h(\theta, T(x_t, x_{t-1})) $, for some functions $g(\cdot)$, $h(\cdot)$ and $T(\cdot)$.
\end{defn}

\reminder{alternative definition}
\begin{defn}
A transition model $p(x_t \given x_{t-1}, \theta)$ is \em{separable} if it can be written as $p(x_t \given x_{t-1}, \theta) =  \exp( g(\theta) \cdot T(x_{t-1}, x_t) - A(\theta) + B(x_{t-1}, x_t) )$, for some functions $g(\cdot)$, $T(\cdot)$, $A(\cdot)$, and $B(\cdot)$.
\end{defn} }

\begin{theorem}
\label{thm:separability}
For a separable system, there exist fixed-dimensional sufficient statistics for the Gibbs density, $p(\theta \mid x_{0:T})$.
\end{theorem}
The proof is straightforward by the Fisher--Neyman factorization theorem; 
more details are given in the supplementary material of the full version \cite{erol2013extended}.

\hide{
\begin{theorem}
\label{thm:separability}
In state-space models defined in \eqref{eq:state-space-model}, if the transition function is {\em separable} and $\theta$ is only involved in transition, there exists sufficient statistics for the Gibbs density, $p(\theta \mid x_{0:T},y_{0:T})$.
\end{theorem} }

The Gaussian system process models defined in \eqref{eq:storvik_model}
are separable, since the transition function
$\mymat{F}_t^T\theta=(\myvec{F}_t)^T\theta$, but the property---and
therefore Storvik's algorithm---applies to a much broader class of
systems.  Moreover, as we now show, non-separable systems may in some
cases be well-approximated by separable systems, constructed by polynomial
density approximation steps
applied to either the Gibbs distribution
$p(\theta \mid x_{0:t})$ or to the transition model.


\section{The extended parameter filter}
\label{sec:algorithm}
Let us consider the following model.
\beq
	x_t=f_{\theta}(x_{t-1})+v_t; \> v_t \sim N(0,\Sigma)
\eeq
where $x \in \mathbb{R}^d$,$\theta \in \mathbb{R}^p$ and $f_{\theta}(\cdot):
\mathbb{R}^d \rightarrow \mathbb{R}^d$ is a vector-valued function parameterized
by $\theta$. We assume that the transition function $f_{\theta}$ may be non-separable. Our
algorithm will create a polynomial approximation to either the transition
function or to the Gibbs distribution, $p(\theta \mid x_{0:t})$. 

To illustrate, let us consider the transition model
$f_{\theta}(x_{t-1})=\sin(\theta x_{t-1})$. It is apparent that this transition
model is non-separable. If we approximate the transition function with a
Taylor series in $\theta$ centered around zero
\beq
	f_{\theta}(x_{t-1}) \approx \hat{f}_{\theta}(x_{t-1}) = x_{t-1}\theta - \frac{1}{3!}x_{t-1}^3\theta^3+ \dots
\eeq
and use $\hat{f}$ as an approximate transition model, the system will become separable.
Then, Storvik's filter can be applied in constant time per update. This Taylor
approximation leads to a log-polynomial density of the form of \eqref{eq:exponential-form}.

Our approach is analogous to that of the extended Kalman filter (EKF). EKF
linearizes nonlinear transitions around the current estimates of the mean and
covariance and uses Kalman filter updates
for state estimation \citep{welch1995introduction}. Our proposed algorithm, which
we call the extended parameter filter (EPF), approximates
a non-separable system with a separable one, using a polynomial
approximation of some arbitrary order. This separable, approximate model is
well-suited for Storvik's filter and allows for constant time updates to the Gibbs density of the
parameters. 

Although we have described an analogy to the EKF, it is important to note that
the EPF can effectively use higher-order approximations instead of just first-order
linearizations as in EKF. In EKF, higher order approximations lead to
intractable integrals. The prediction integral for EKF
\beq
	p(x_t \mid y_{0:t-1} ) =\int p(x_{t-1} \mid y_{0:t-1})p(x_t \mid x_{t-1})dx_{t-1} \nn
\eeq
can be calculated for linear Gaussian transitions, in which case the mean and the covariance matrix
are the tracked sufficient statistic. However, in the case of quadratic
transitions (or any higher-order transitions), the above
integral is no longer analytically tractable. 

In the case of EPF, the transition model is the identity transition
and hence the prediction step is trivial. The filtering recursion is 
\hide{densities that grow in size with time (i.e., $p(x_t \mid y_{0:t})$ will not be a
density that is proportional to $p(x_{t-1} \mid y_{0:t-1})$ due to the integral
required for prediction). We describe how EPF avoids this growth in density
size. Suppose we assume an identity transition model,
$p(x_t \mid x_{t-1})=\delta(x_t - x_{t-1})$. Then the Bayesian recursions defined by equation \ref{eq:bayes1} will lead to
\beq
	p(X_t=x_t \mid y_{0:t}) \propto p(y_t \mid x_t)p(X_{t-1}=x_t \mid y_{0:t-1})
\eeq 
which does not grow in size. In the static parameter estimation case, this corresponds to }
\beq
\label{parameter_recursions}
	p(\theta \mid x_{0:t}) \propto p(x_t \mid x_{t-1},\theta) p(\theta \mid
  x_{0:t-1}).
\eeq
We approximate the transition $p(x_t \mid x_{t-1},\theta)$ with a log-polynomial
density $\hat{p}$ (log-polynomial in $\theta$), so that the Gibbs density, which 
satisfies the recursions in equation \ref{parameter_recursions}, has a fixed
log-polynomial structure at each time step. Due to the polynomial structure, the
approximate Gibbs density can be tracked in terms of its sufficient statistic
(i.e., in terms of the coefficients of
the polynomial). The log-polynomial structure is derived in 
\secref{sec:online_gibbs_approx}. Pseudo-code for EPF is shown in Algorithm 3.
\begin{algorithm2e}[tb]
\label{alg:EPF}
\caption{Extended Parameter Filter}
\KwResult{Approximate the Gibbs density  $p(\theta \mid x_{0:t},y_{0:t})$ with the log-polynomial density $\hat{p}(\theta \mid x_{0:t},y_{0:t})$}
\KwOut{$\tilde{x}^1 \dots \tilde{x}^N$}
initialize $\left\{x_0^i\right\}$ and $S_0^i \leftarrow 0$\; 
\For{$t=1,\ldots,T$} {
\For{$i=1,\dots,N$}{
$S_t^i =  \update(S_{t-1}^i, x_{t-1})$ \tcp*{
update statistics for polynomial approximation $\log(\hat{p}(\theta | \bar{x}_{0:t-1}, y_{0:t-1}))$}
	sample $\theta^i \sim \hat{p}(\theta \mid \myvec{\bar{x}}_{0:t-1}^i,\myvec{y}_{0:t-1})=\hat{p}(\theta \mid {S}^i_t)$ \;
	sample $\myvec{x}_{t}^i \sim p(\myvec{x_t} \mid \myvec{\bar{x}}_{t-1}^i,\theta^i)$ \;
	$w^i \leftarrow p(\myvec{y}_t \mid \myvec{x}_t^i,\theta^i)$\;
}
sample $\left\{\frac{1}{N},\bar{{x}}_t^i,\bar{{S}}_t^i \right\}\leftarrow  $Multinomial$\left\{w_t^i,{x}_t^i, S_t^i \right\}$\;
$\left\{{x}_t^i, S_t^i \right\}\leftarrow \left\{\bar{{x}}_t^i, \bar{{S}}_t^i\right\}$\;
}
\end{algorithm2e}

Note that the approximated Gibbs density will be a
log-multivariate polynomial density of fixed order (proportional to the order of
the polynomial approximation). Sampling from such a density is not
straightforward but can be done by Monte Carlo sampling. We suggest slice
sampling \citep{neal2003slice} or the Metropolis-Hastings algorithm
\citep{robert2005monte} for this purpose. Although some approximate sampling
scheme is necessary, sampling from the approximated density remains a constant-time operation when the dimension of  $\hat{p}$ remains constant. 

It is also important to note that performing a polynomial approximation for a $p$-dimensional parameter space may not be an easy task. However, we can reduce the
computational complexity of such approximations by exploiting locality
properties. For instance, if
$f_{\theta}(\cdot)=h_{\theta_1,\dots,\theta_{p-1}}(\cdot)+g_{\theta_p}(\cdot)$,
where $h$ is separable and $g$ is non-separable, we only need to approximate $g$.

In section \ref{sec:poly_general}, we discuss the validity of the
approximation in terms of the KL-divergence between the true and approximate
densities. In section \ref{sec:poly_static}, we analyze the distance between
an arbitrary density and its approximate form with respect to the order of the
polynomial. We show that the distance goes to zero
\textit{super-exponentially}. Section \ref{sec:online_gibbs_approx} analyzes
the error for the static parameter estimation problem and introduces the
form of the log-polynomial approximation.

\section{Approximating the conditional distribution of parameters}
\label{sec:poly_general}
%
%
%
In this section, we construct approximate sufficient statistics for arbitrary one--dimensional state space models.
We do so by exploiting log-polynomial approximations to arbitrary probability densities. We prove that such approximations can be made arbitrarily accurate. Then, we
analyze the error introduced by log-polynomial approximation for the arbitrary one--dimensional model.
\subsection{Taylor approximation to an arbitrary density}
Let us assume a distribution $p$ (known only up to a normalization constant)
expressed in the form $p(x) \propto \exp(S(x))$, where $S(x)$ is an analytic
function on the support of the distribution. In general we need a Monte Carlo
method to sample from this arbitrary density. In this section, we describe an
alternative, simpler sampling method.
We propose that with a polynomial approximation $P(x)$ (Taylor, Chebyshev etc.) of
sufficient order to the function $S(x)$, we may sample from a distribution
$\widehat{p} \propto \exp(P(x))$  with  a simpler (i.e. log-polynomial) structure. We 
show that the distance between the distributions $p$ and $\widehat{p}$ reduces
to $0$ as the order of the approximation increases. 

The following theorem is based on Taylor 
approximations; however, the theorem can be generalized to handle any polynomial approximation scheme. The proof is given in \cite{erol2013extended}.

\begin{theorem}
Let $S(x)$ be a $M+1$ times differentiable function with bounded derivatives, and
let $P(x)$ be its $M$-th order Taylor approximation. Then the KL-divergence between
distributions $p$ and $\hat{p}$ converges to $0$, \textbf{super-exponentially}
as the order of approximation $M \to \infty$.
\label{thm:kl_converge}
\end{theorem}

\begin{figure}[tb]
\centering
    \includegraphics[width=2in]{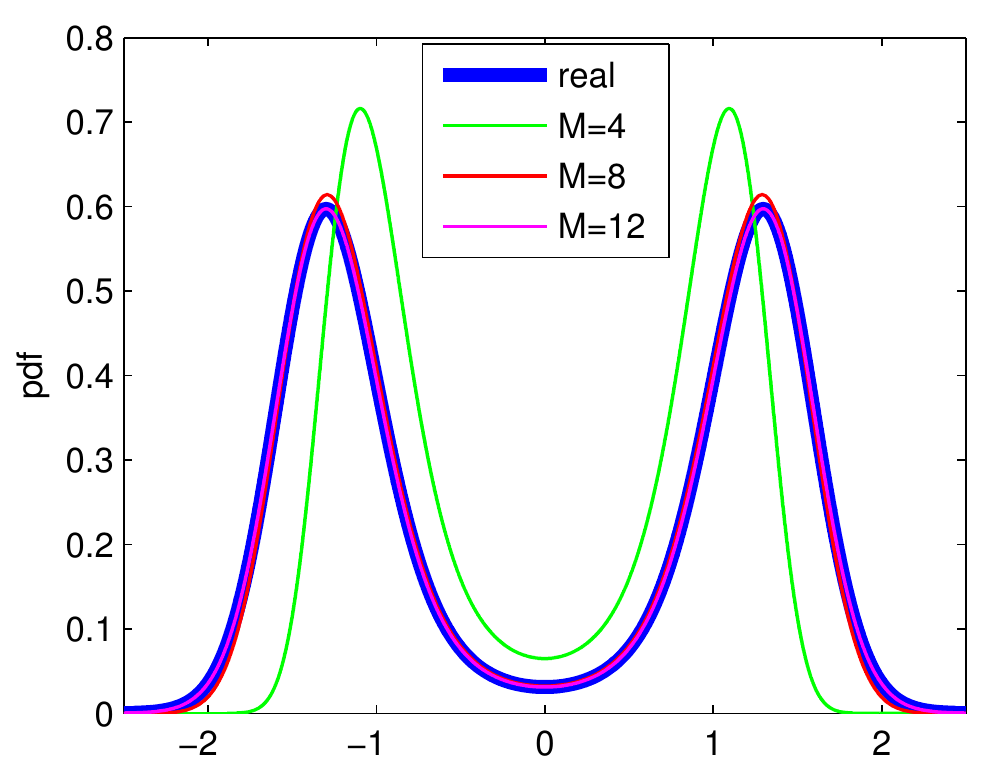}
    \vspace{-0.1in}
\caption{Approximated PDFs to the order $M$.}
\label{fig:pdfs}
\end{figure}

We validate the Taylor approximation approach for the log-density
$S(x)= -x^2 +5\sin^2(x)$. Figure \ref{fig:pdfs} shows the result for this case.

\label{sec:poly_static}
\subsection{Online approximation of the Gibbs density of the parameter }
\label{sec:online_gibbs_approx}
In our analysis, we will assume the following model. 
\begin{subequations}
\label{eq:model}
\begin{align*}
x_t &= f_{\theta}(x_{t-1})+v_t, \>v_t \sim N(0,\sigma^2) \nn \\
y_t &= g(x_t)+w_t,\> w_t \sim N(0,\sigma_{o}^2)
\end{align*}
\end{subequations}
The posterior distribution for the static parameter is 
\[p(\theta | x_{0:T}) \propto p(\theta) \prod_{t=1}^T p(x_t | x_{t-1},\theta).\]
The product term, which requires linear time, is the bottleneck for this
computation. A polynomial approximation to the transition function
$f_{\theta}(\cdot)$ (the Taylor approximation around $\theta=0$) is:
{\small
\begin{align*}
f_{\theta}(x_{t-1}) &= h(x_{t-1},\theta) 
= \sum_{i=0}^M
\underbrace{\frac{1}{i!}\frac{d^ih(x_{t-1},\theta^i)}
{d\theta}\big|_{\theta=0}}_{H^i(x_{t-1})}\theta^i + R_M(\theta) \\
&= \sum_{i=0}^{M}H^i(x_{t-1})\theta^i + R_M(\theta) 
= \hat{f}(\theta) + R_M(\theta)
\end{align*}
}
where $R_M$ is the error for the $M$-dimensional Taylor approximation. We define coefficients $J^i_{x_{t-1}}$ to satisfy 
$\left(\sum_{i=0}^{M}H^i(x_{t-1})\theta^i\right)^2=J^{2M}_{x_{t-1}}\theta^{2M}+\dots+J^{0}_{x_{t-1}}\theta^{0}$.

Let $\hat{p}(\theta\mid x_{0:T})$ denote the approximation to $p(\theta \mid
x_{0:T})$ obtained by using the polynomial approximation to $f_\theta$
introduced above. 

\begin{theorem}
\label{thm:exp_family}
$\hat{p}(\theta\mid x_{0:T})$ is in the exponential family with the log-polynomial density
{\small
\begin{align}
\label{eq:exponential-form}
	&\log p(\theta)+ \\
	&\underbrace{ \begin{pmatrix}
		\theta^1\\ 
		\vdots\\ 
		\theta^M\\ 
		\theta^{M+1} \\
		\vdots\\ 
		\theta^{2M}
	\end{pmatrix}^T }_{T(\theta)^T}  .  
	\underbrace{ \begin{pmatrix}
		\frac{1}{\sigma^2}\sum_{k=1}^T x_k H^1(x_{k-1})-\frac{1}{2\sigma^2}\sum_{k=1}^T J^1_{x_{k-1}}\\ 
		\vdots\\ 
		\frac{1}{\sigma^2}\sum_{k=1}^T x_k H^M(x_{k-1})-\frac{1}{2\sigma^2}\sum_{k=1}^T J^M_{x_{k-1}}\\ 
		-\frac{1}{2\sigma^2}\sum_{k=1}^T J^{M+1}_{x_{k-1}} \\
		\vdots\\
		-\frac{1}{2\sigma^2}\sum_{k=1}^T J^{2M}_{x_{k-1}}
	\end{pmatrix}   }_{\eta(x_0,\dots,x_t)} \notag 
\end{align}
}
\end{theorem}
The proof is given in the supplementary material.

This form has finite dimensional sufficient statistics. Standard sampling from $p(\theta \mid x_{0:t})$ requires $O(t)$ time, whereas
with the polynomial approximation we can sample from this structured
density of fixed dimension in constant time (given that sufficient statistics were tracked). We can furthermore prove that
sampling from this exponential form approximation is asymptotically correct.

\begin{theorem}
\label{thm:static_error}
Let $p_T(\theta \mid x_{0:T})$ denote
the Gibbs distribution and $\widehat{p}_T(\theta \mid x_{0:T})$ its order $M$
exponential family approximation. Assume that parameter $\theta$ has support
$\mathcal{S}_{\theta}$ and finite variance. Then as $M \to \infty, T \to
\infty$, the KL divergence between $p_T$ and $\widehat{p}_T$ goes to zero.
\[\lim_{M, T \to \infty} D_{KL}(p_T \mid \mid \hat{p}_T) = 0 \]
\end{theorem}
The proof is given in the supplementary material \cite{erol2013extended}.
Note that
the analysis above can be generalized to higher dimensional parameters. The one
dimensional case is discussed for ease of exposition.

In the general case, an order $M$ Taylor expansion for a $p$ dimensional
parameter vector $\theta$ will have $M^p$ terms. Then each update of the
sufficient statistics will cost $O(M^p)$ per particle, per time step, yielding the
total complexity $O(NTM^p)$. 
However, as noted before, we can often exploit
the local structure of $f_\theta$ to speed up the update step. Notice that in
either case, the update cost per time step is fixed (independent of $T$).

\section{Experiments}
\label{sec:experiment}
\begin{figure}[tb]
\centering
    \includegraphics[width=2in]{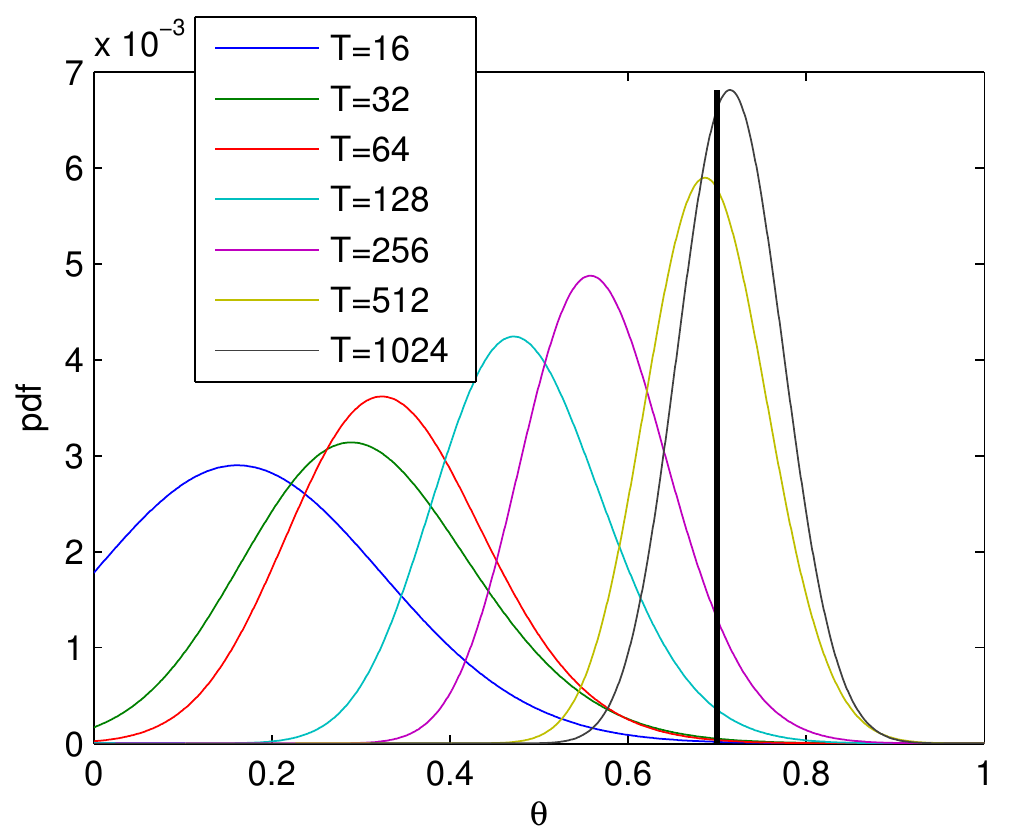}
\vspace{-0.1in}
\caption{Sinusoidal dynamical model (SIN). Shrinkage of the Gibbs density
$p(\theta \mid x_{0:T})$ with respect to time duration $T$. Note that as $T$
grows, the Gibbs density converges to the true parameter value. }
\label{fig:shrinkage}
\end{figure}

\begin{figure}[tb]
\vspace{-0.1in}
\centering
\subfigure[Approximating Gibbs]{
  \includegraphics[width=1.5in]{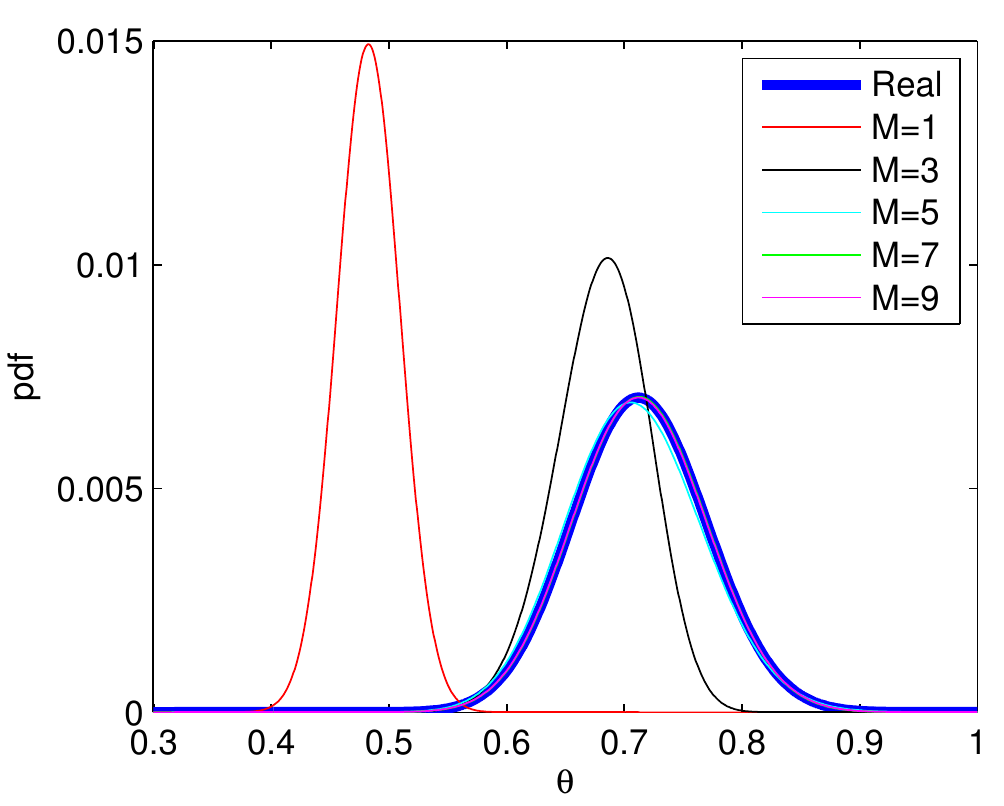}
  \label{fig:convergence_wrt_N}
}
\subfigure[KL-divergence]{
    \includegraphics[width=1.5in]{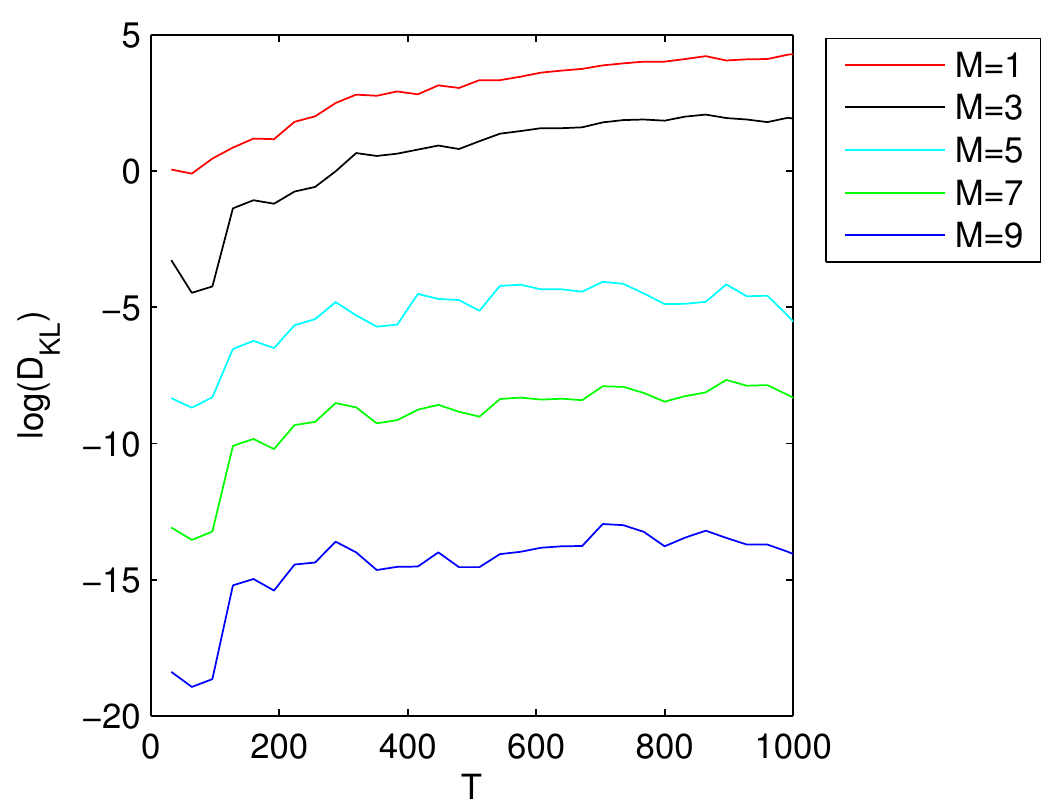}
    \label{fig:KLdiv}
}
\vspace{-0.1in}
\caption{Sinusoidal dynamical model (SIN). 
\subref{fig:convergence_wrt_N} Convergence of the approximate densities to the
Gibbs density $p(\theta \mid x_{0:1024})$ with respect to the approximation
order $M$; \subref{fig:KLdiv}  KL-divergence $D_{KL}(p \mid \hat{p})$ with
respect to duration $T$ and approximation order $M$.}
\end{figure}

\begin{figure*}[tb]
\vspace{-0.1in}
\centering
\subfigure[Particle filter (SIR)]{
    \includegraphics[width=2in]{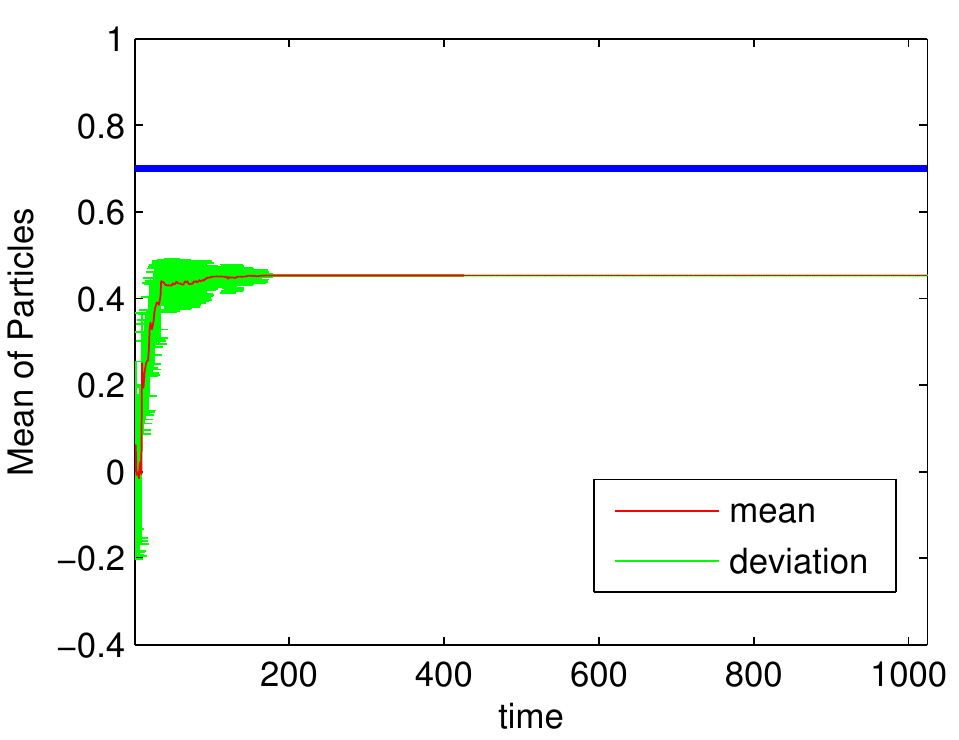}
    \label{fig:pf_collapse}
}
\subfigure[Liu--West filter] {
    \includegraphics[width=2in]{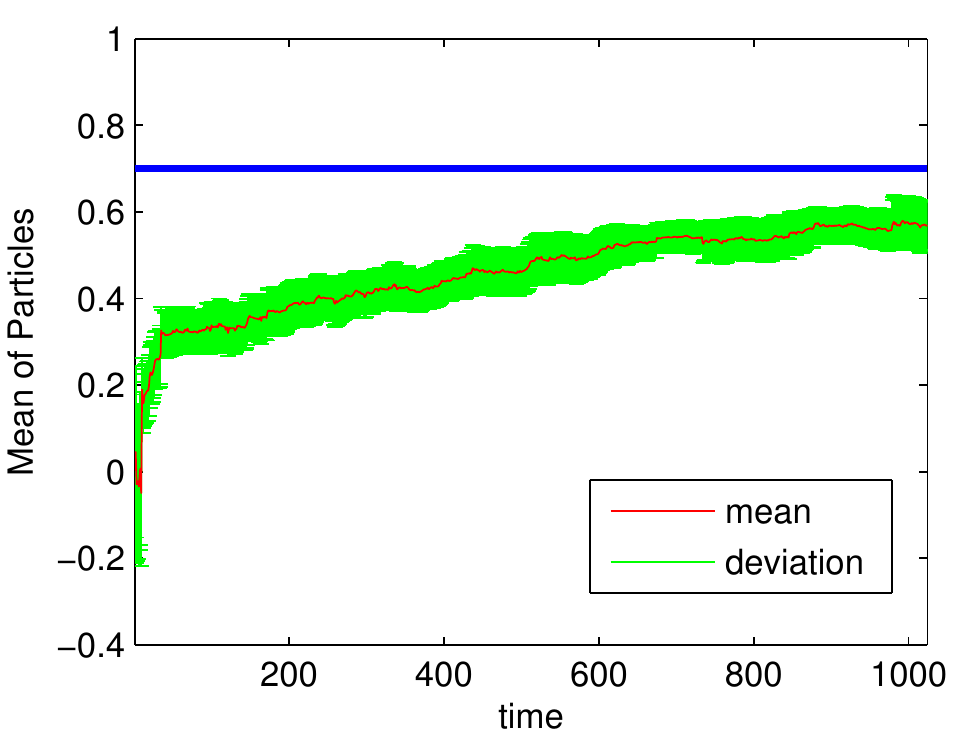}
    \label{fig:liu-west}
}
\subfigure[EPF]{
  \includegraphics[width=2.1in]{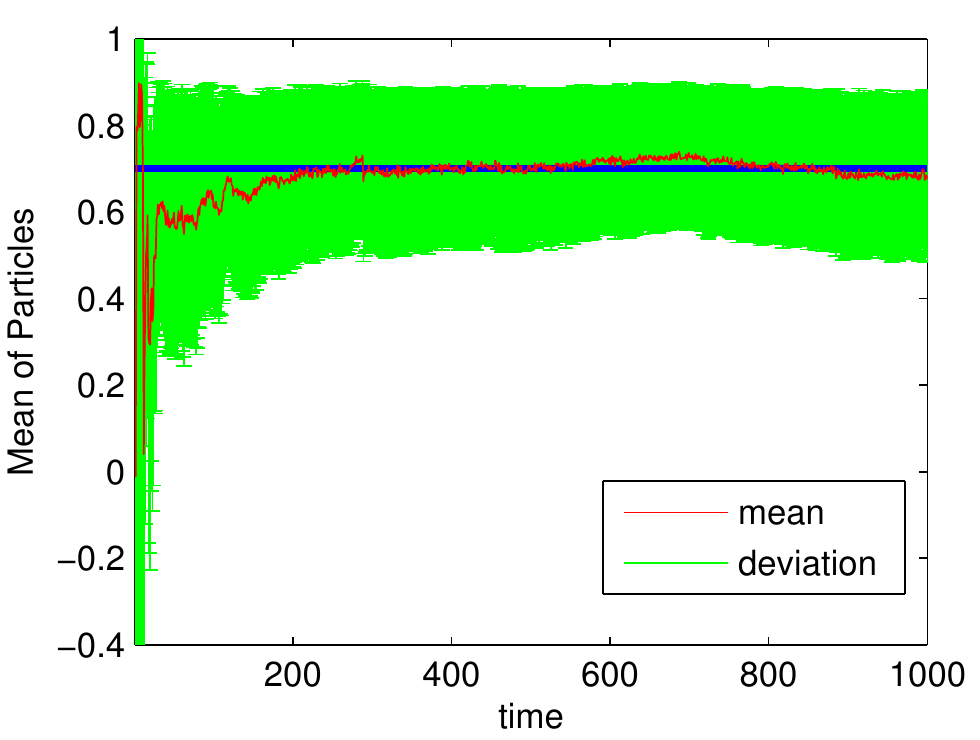}
  \label{fig:storvik1000}
}
\vspace{-0.1in}
\caption{Sinusoidal dynamical model (SIN).
\subref{fig:pf_collapse}: Particle filter (SIR) with $N = 50000$ particles. Note
the failure to converge to the true value of parameter $\theta$ ($0.7$, shown as
the blue line). 
\subref{fig:liu-west}: Liu--West filter with $N = 50000$ particles. 
\subref{fig:storvik1000}: EPF with $N = 1000$ particles and $7$-th order
approximation. Note both SIR and Liu--West do not converge, while the EPF converges quickly even with orders of magnitude fewer particles. }
\end{figure*}

The algorithm is implemented for three specific cases. Note
that the models discussed do not satisfy the Gaussian process model assumption of \citet{storvik2002particle}.
\subsection{Single parameter nonlinear model}
Consider the following model with sinusoid transition dynamics (SIN):
\begin{align}
	x_t&=\sin(\theta x_{t-1})+v_t, \>v_t \sim N(0,\sigma^2) \nn \\
	y_t&=x_t+w_t,\> w_t \sim N(0,\sigma_{\obs}^2)
\end{align}
where $\sigma=1$, $\sigma_{\obs}=0.1$ and the Gaussian prior for parameter
$\theta$ is $N(0,0.2^2)$. The observation sequence is generated by sampling from
SIN with true parameter value $\theta=0.7$. 

Figure \ref{fig:shrinkage} shows how the Gibbs density $p(\theta \mid x_{0:t})$
shrinks with respect to time, hence verifying identifiability for this model.
Notice that as $T$ grows, the densities concentrate around the true
parameter value. 

A Taylor approximation around $\theta=0$ has been applied to the transition
function $\sin(\theta x_t)$. Figure \ref{fig:convergence_wrt_N} shows the
approximate densities for different polynomial orders for $T=1024$. Notice that
as the polynomial order increases, the approximate densities converge to the
true density $p(\theta \mid x_{0:1024})$.

The KL-divergence $D_{KL}(p \mid \mid \hat{p})$ for different polynomial orders (N) and different data lengths (T)
is illustrated in  \figref{fig:KLdiv}. The results are consistent with the theory developed in \secref{sec:poly_static}.

The degeneracy of a bootstrap filter with $N = 50000$ particles can be seen from figure
\ref{fig:pf_collapse}. The Liu--West approach with $N = 50000$ particles is shown
in \ref{fig:liu-west}. The perturbation is $\theta_t = \rho \theta_{t-1} +
(1-\rho)\bar{\theta}_{t-1} +\sqrt{1-\rho^2}\std(\theta_{t-1})N(0,1)$, where
$\rho=0.9$. Notice that even with $N = 50000$ particles and large perturbations,
the Liu--West approach converges slowly compared to our method. Furthermore,
for high-dimensional spaces, tuning the perturbation parameter
$\rho$ for Liu--West becomes difficult.

The EPF has been implemented on this model with $N = 1000$ particles with a
$7$-th order Taylor approximation to the posterior. The time complexity is $O(NT)$. The mean and the standard
deviation of the particles are shown in figure \ref{fig:storvik1000}.

\subsection{Cauchy dynamical system}
We consider the following model.
\begin{align}
x_t &= a x_{t-1} + \Cauchy(0,\gamma) \\
y_t &= x_t + N(0,\sigma_{\obs}) 
\end{align}
Here $\Cauchy$ is the Cauchy distribution centered at $0$ and with shape
parameter $\gamma=1$. We use $a = 0.7$, $\sigma_{\obs} = 10$, where the prior
for the AR(1) parameter is $N(0,0.2^2)$. This model
represents autoregressive time evolution with heavy-tailed noise. 
Such heavy-tailed noises are observed in network traffic data and click-stream data.
The standard Cauchy distribution we use is
\beq
	f_v(v;0,1)=\frac{1}{\pi(1+v^2)} = \exp\left(-\log(\pi) -\log(1+v^2)\right). \nn
\eeq
We approximate $\log(1+v^2)$ by $v^2-v^4/2+v^6/3 -v^8/4
+\dots$ (the Taylor approximation at 0).

\begin{figure*}[tb]
\centering
\vspace{-0.1in}
\subfigure[Data sequence]{        \includegraphics[width=1.55in]{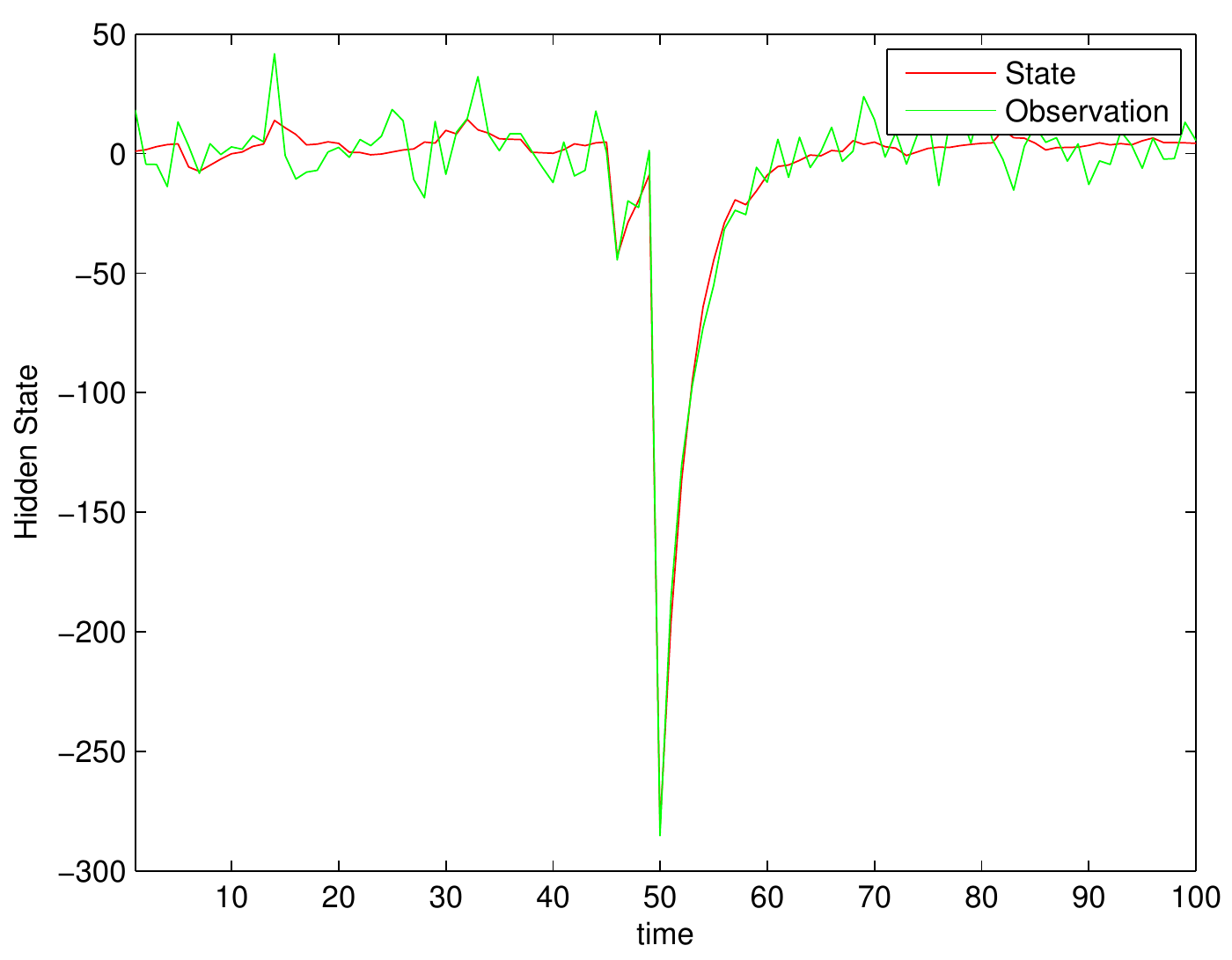}
 \label{fig:cauchy_process}
}
\subfigure[Particle filter (SIR)]{
\includegraphics[width=1.55in]{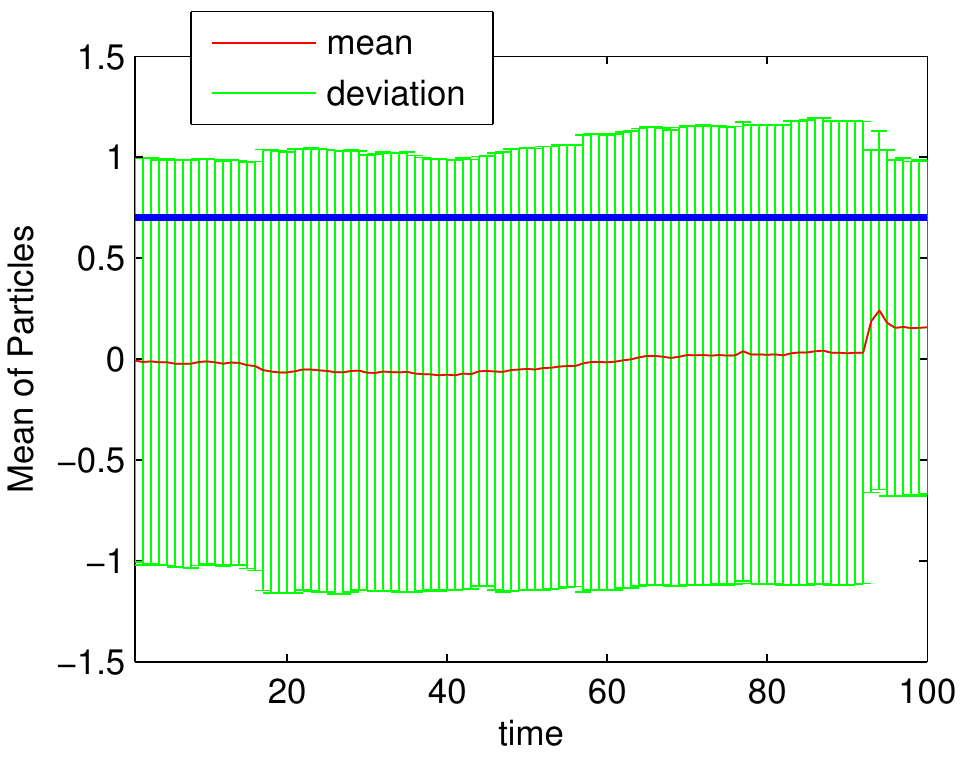}
\label{fig:pf50000cauchy}
}
\subfigure[Liu--West filter]{
    \includegraphics[width=1.55in]{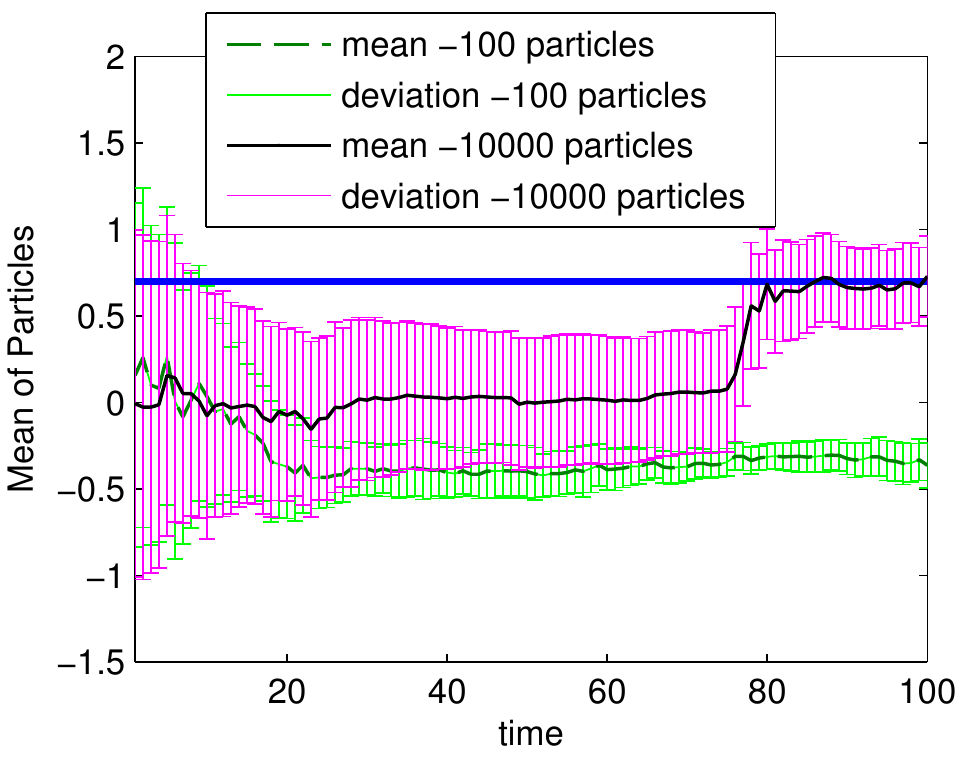}
\label{fig:liu_west_cauchy}
}
\subfigure[EPF]{
    \includegraphics[width=1.55in]{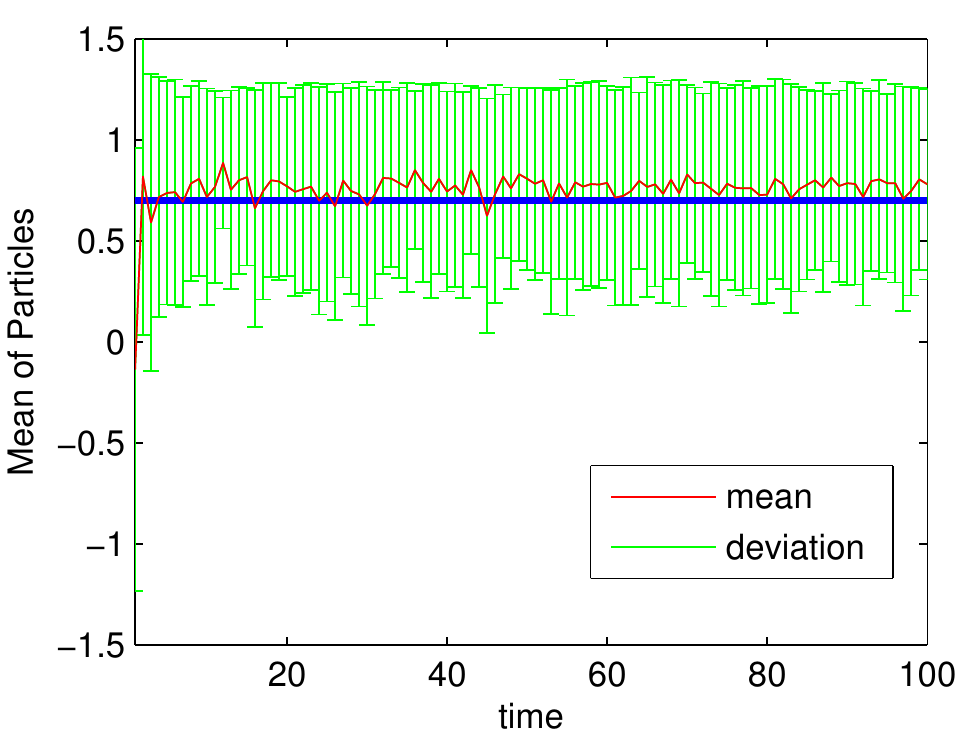}
\label{fig:storvik100cauchy}
}
\vspace{-0.1in}
\caption{Cauchy dynamical system.  \subref{fig:cauchy_process}: Example sequences for  hidden states and observations. 
\subref{fig:pf50000cauchy}: Particle filter estimate with 50000 particles. 
\subref{fig:liu_west_cauchy}: Liu--West filter with 100 and 10000 particles. 
\subref{fig:storvik100cauchy}: EPF using only 100 particles and 10th order approximation. Note EPF converges to the actual value of parameter $a$ (=0.7, in blue line) while SIR does not even with orders of magnitude more particles, neither does Liu--West with the same number of particles.}
\end{figure*}

Figure \ref{fig:cauchy_process} shows the simulated hidden state and the
observations ($\sigma_{obs}=10$).  Notice that the simulated process differs
substantially from a standard AR(1) process due to the heavy-tailed noise. Storvik's filter cannot handle this model since the
necessary sufficient statistics do not exist.

Figure \ref{fig:pf50000cauchy} displays the mean value estimated by a bootstrap
filter with $N = 50000$ particles. As before the bootstrap filter is unable to
perform meaningful inference.  Figure \ref{fig:liu_west_cauchy} shows the
performance of the Liu--West filter with both $N = 100$ and $N = 10000$
particles. The Liu--West filter does not converge for $N = 100$ particles and
converges slowly for $N = 10000$ particles. Figure \ref{fig:storvik100cauchy}
demonstrates the rapid convergence of the EPF for only $N = 100$ particles with 10th order approximation. The time complexity is $O(NT)$.

Our empirical results confirm that the EPF proves useful for models with
heavy-tailed stochastic perturbations.

\subsection{Smooth Transition AR model}

\begin{figure*}[tb]
\centering
\vspace{-0.1in}
\subfigure[Gibbs density]{
    \includegraphics[width=1.72in]{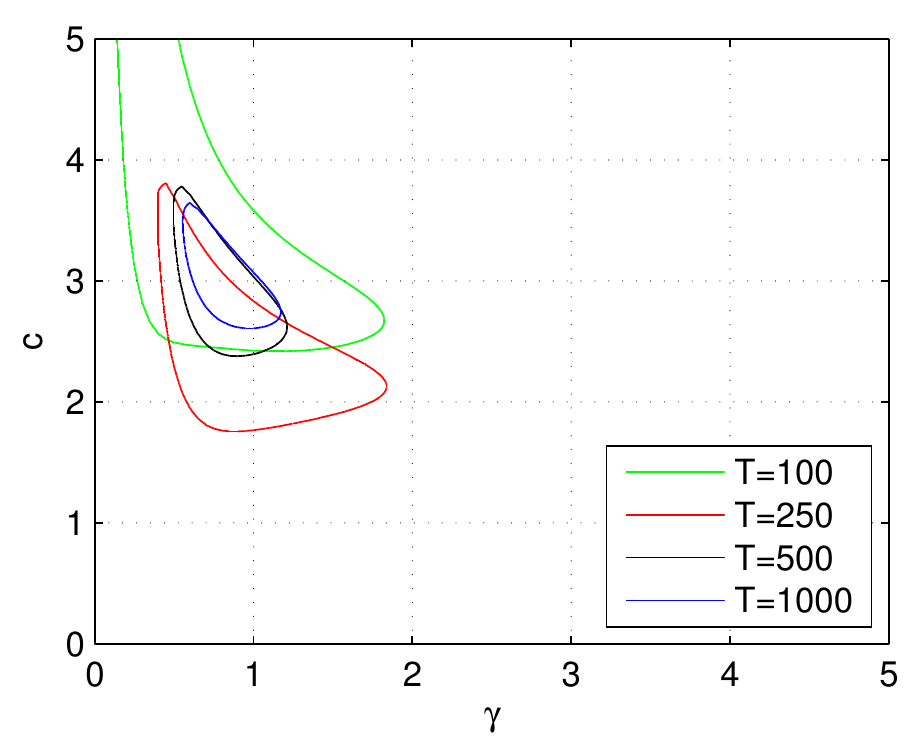}
    \label{fig:star_shrink}
}
\subfigure[Liu--West filter]{
    \includegraphics[width=1.8in]{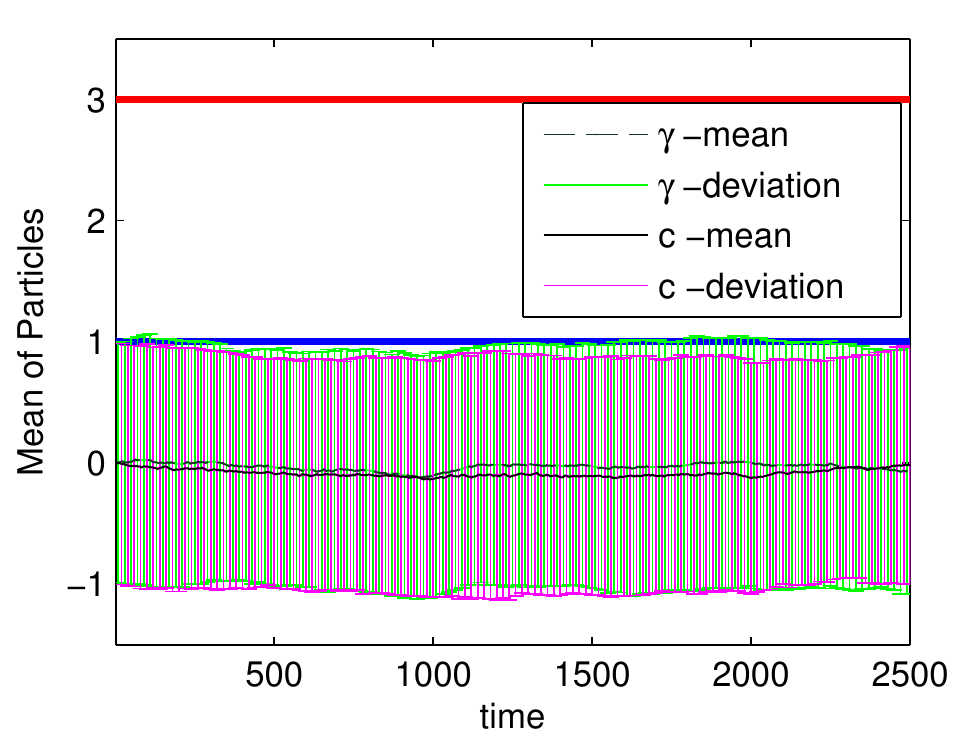}
    \label{fig:liu_west_star}
}
\subfigure[EPF]{
    \includegraphics[width=1.8in]{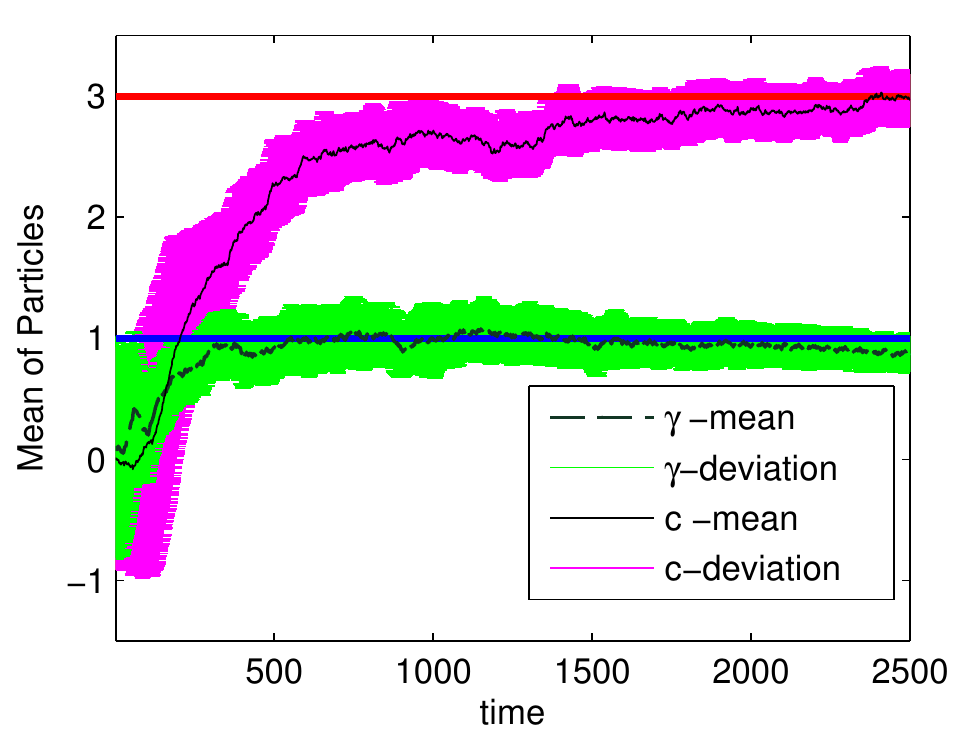}
    \label{fig:storvik_2d_100}
}
\vspace{-0.1in}
\caption{STAR model. \subref{fig:star_shrink}: Shrinkage of the Gibbs density $p(\gamma,c \mid x_{0:t})$ with respect to time.
\subref{fig:liu_west_star}: Liu--West filter using 50000 particles.
\subref{fig:storvik_2d_100}: 
EPF using 100 particles and 9th order approximation. Note the EPF's estimates for both parameters converge to the actual values quickly even with only 100 particles, while Liu--West does not converge at all.
}
\end{figure*}
The smooth transition AR (STAR) model is a smooth generalization of the self-exciting threshold autoregressive (SETAR) model, \cite{star_model}. It is generally expressed in the following form.
\begin{align}
	x_t =& \left( a_1 x_{t-1} +a_2 x_{t-2} + \dots + a_p x_{t-p} \right) \left[ 1- G(x_{t-d}; \gamma,c) \right ] &  \nn \\
		+ &\left( b_1 x_{t-1} +b_2 x_{t-2} + \dots + b_p x_{t-p} \right) \left[  G(x_{t-d}; \gamma,c) \right ] + \epsilon_t  & \nn
\end{align}
where $\epsilon_t$ is i.i.d. Gaussian with mean zero and variance $\sigma^2$ and $G(\cdot)$ is a nonlinear function of $x_{t-d}$, where $d >0$. We will use the logistic function 
\beq
	G(y_{t-d};\gamma,c)= \frac{1} { 1+ \exp\left(-\gamma (x_{t-d}- c) \right) }
\eeq
For high $\gamma$ values, the logistic function converges to the indicator
function, $\mathbb{I}(x_{t-d} >c)$, forcing STAR to converge to SETAR (SETAR
corresponds to a switching linear--Gaussian system). We will use $p=1=d$,  where
$a_1=0.9$ and $b_1=0.1$ and $\sigma=1$ (corresponding to two different AR(1)
processes with high and low memory). We attempt to estimate parameters $\gamma,
c$ of the logistic function, which have true values $\gamma=1$ and $c=3$. Data  (of
length $T=1000$) is generated from the model under fixed parameter values and with 
observation model $y_t=x_t+w_t$, where $w_t$ is additive Gaussian noise with
mean zero and standard deviation $\sigma_{\obs}=0.1$. Figure \ref{fig:star_shrink} shows the shrinkage of the Gibbs density $p(\gamma,c \mid x_{0:T})$, verifying identifiability. 

The non-separable logistic term is approximated as
\begin{align*}
&	\frac{1}{1+\exp\left(-\gamma(x_{t-1}-c) \right)} \\
&		\approx \frac{1}{2} - \frac{1}{4} \gamma(c-x_{t-1}) + \frac{1}{48}\gamma^3 (c-x_{t-1})^3 +\dots  \nn
\end{align*}

Figure \ref{fig:liu_west_star} displays the failure of the Liu--West
filter for $N = 50000$ particles.  Figure \ref{fig:storvik_2d_100} shows the
mean values for $\gamma, c$ from EPF for only $N=100$ particles with $9$th order
Taylor approximation. Sampling from the log-polynomial approximate density is
done through the random-walk Metropolis--Hastings algorithm. For each particle
path, at each time step $t$, the Metropolis--Hastings sampler is initialized from the
parameter values at $t-1$. The burn-in period is set to be 0, so only one MH
step is taken per time step (i.e., if a proposed sample is more likely it is
accepted, else it is rejected with a specific probability). The whole filter has time complexity $O(NT)$.

\section{Conclusion}
\label{sec:conclusion}
Learning the parameters of temporal probability models remains a significant open problem for practical applications.
We have proposed the extended parameter filter (EPF), 
a novel approximate inference algorithm that combines Gibbs sampling of parameters with computation of approximate sufficient statistics.
The update time for EPF is independent of the length of the observation
sequence. Moreover, the algorithm has provable error bounds and handles a wide variety of models. Our experiments confirm these properties and illustrate difficult cases on which EPF works well.

One limitation of our algorithm is the complexity of Taylor
approximation for high-dimensional parameter vectors. 
We noted that, in some cases, the process can be decomposed into lower-dimensional subproblems.
Automating this step would be beneficial.

\bibliographystyle{plainnat}
\bibliography{BIB/leiliref}

\twocolumn[
  \icmltitle{The Extended Parameter Filter}
]
\appendixpage
\appendix
\section{Storvik's filter as a Kalman filter }
\label{sec:appendix2}
Let us consider the following model.
\begin{align}
	\myvec{x}_t&=\mymat{A}\myvec{x}_{t-1}+\myvec{v}_t, \> \myvec{v}_t \sim N(0,\mymat{Q}) \nn \\
	\myvec{y}_t&=\mymat{H}\myvec{x}_t+\myvec{w}_t, \> \myvec{w}_t \sim N(0,\mymat{R})
\end{align}
We will call the MMSE estimate Kalman filter returns as $\myvec{x}_{t \mid t}=\mathbb{E}[\myvec{x}_t \mid \myvec{y}_{0:t}]$ and the variance $\mymat{P}_{t \mid t}=cov(\myvec{x}_t \mid \myvec{y}_{0:t})$. Then the update for the conditional mean estimate is as follows.
\begin{align}
	& \myvec{x}_{t \mid t}  = \mymat{A}\myvec{x}_{t-1 \mid t-1} \nn \\
	&+ \underbrace{ \mymat{P}_{t \mid t-1}\mymat{H}^T(\mymat{H}\mymat{P}_{t \mid t-1} \mymat{H}^T+ \mymat{R})^{-1} }_{\mymat{K}_t}(\myvec{y}_t-\mymat{H}\mymat{A}\myvec{x}_{t-1 \mid t-1}) \nn 
\end{align}
where as for the estimation covariance
\begin{align}
	\mymat{P}_{t \mid t-1} &= \mymat{A}\mymat{P}_{t-1 \mid t-1} \mymat{A}^T + \mymat{Q} \nn \\
	\mymat{P}_{t \mid t}     &= ( \mymat{I}-\mymat{K}_t\mymat{H}) \mymat{P}_{t \mid t-1}
\end{align}
Matching the terms above to the updates in equation \ref{eq:storvik_updates}, one will obtain a linear model for which the transition matrix is $\mymat{A}=\mymat{I}$, the observation matrix is $\mymat{H}=\mymat{F}_t$, the state noise covariance matrix is $\mymat{Q}=\mymat{0}$, and the observation noise covariance matrix is $\mymat{R}=\mymat{Q}$

\section{Proof of theorem \ref{thm:separability} }
\label{sec:appendix3}
Let us assume that $\myvec{x} \in \mathbb{R}^d$,$\theta \in \mathbb{R}^p$ and $f_{\theta}(\cdot): \mathbb{R}^d \rightarrow \mathbb{R}^d$ is a vector valued function parameterized by $\theta$. Moreover, due to the assumption of separability $f_{\theta}(\myvec{x}_{t-1})=l(\myvec{x}_{t-1})^Th(\theta)$, where we assume that $l(\cdot): \mathbb{R}^d \rightarrow \mathbb{R}^{m \times d}$ and $h(\cdot): \mathbb{R}^p \rightarrow \mathbb{R}^m$ and $m$ is an arbitrary constant. The stochastic perturbance will have the log-polynomial density $p(\myvec{v}_t) \propto \exp(\mymat{\Lambda}_1 \myvec{v}_t + \myvec{v}_t^T \mymat{\Lambda}_2 \myvec{v}_t + \dots)$ Let us analyze the case of $p(\myvec{v}_t) \propto \exp(\mymat{\Lambda}_1 \myvec{v}_t + \myvec{v}_t^T \mymat{\Lambda}_2 \myvec{v}_t )$, for mathematical simplicity.
\begin{proof}
\begin{align*}
	&\log p(\theta \mid \myvec{x}_{0:T}) \propto \log p(\theta) +\sum_{t=1}^T \log p(\myvec{x}_t \mid \myvec{x}_{t-1},\theta)  \\
	 &\propto \log p(\theta) +\sum_{t=1}^T \mymat{\Lambda}_1 \left(\myvec{x}_t-l(\myvec{x}_{t-1})^Th(\theta)\right) + \\  
	 & \> \>\>\left(\myvec{x}_t-l(\myvec{x}_{t-1})^Th(\theta)\right)^T\mymat{\Lambda}_2 \left(\myvec{x}_t-l(\myvec{x}_{t-1})^Th(\theta)\right)    \\
	 &\propto  \log p(\theta) + \underbrace{\left( \sum_{t=1}^T -(\mymat{\Lambda}_1+2\myvec{x}_t^T\mymat{\Lambda}_2)l(\myvec{x}_{t-1})^T \right)}_{\mymat{S}_1}h(\theta) \\ 
	 \hide{& \trace\left\{ \left( \sum_{i=2}^T g(\myvec{x}_{t-1})\mymat{\Lambda}_2 g(\myvec{x}_{t-1})^T \right) h(\theta)h(\theta)^T \right\} }
	 & + h^T(\theta) \underbrace{ \left(\sum_{t=1}^T l(x_{t-1}) \mymat{\Lambda}_2 l^T(x_{t-1})\right) }_{\mymat{S}_2}h(\theta) + \mathrm{constants}
\end{align*}
Therefore, sufficient statistics ($\mymat{S}_1 \in \mathbb{R}^{1\times m}$ and $\mymat{S}_2 \in \mathbb{R}^{m \times m}$) exist. The analysis can be generalized for higher-order terms in $ \myvec{v}_t$ in similar fashion.
\end{proof}

\section{Proof of theorem \ref{thm:kl_converge} }
\label{sec:appendix1}
\begin{proposition}
Let $S(x)$ be a $M+1$ times differentiable function and $P(x)$ its order $M$
Taylor approximation. Let $I = (x - a, x + a)$ be an open interval around
$x$. Let $R(x)$ be the remainder function, so that $S(x) = P(x) + R(x)$. Suppose
there exists constant $U$ such that

\[ \forall y \in I, \quad \left |f^{(k+1)}(y)\right | \leq  U\]

We may then bound 

\[\forall y \in I, \quad \left | R(y) \right | \leq  U \frac{a^{M+1}}{(M+1)!} \]
 \end{proposition}

We define the following terms

\begin{align*}
\epsilon &= U \frac{a^{M+1}}{(M+1)!} \\
Z &=\int_{I} \exp(S(x))dx \\
\hat{Z} &=\int_{I} \exp(P(x))dx
\end{align*}

Since $\exp(\cdot)$ is monotone and increasing and $\left |S(x) - P(x) \right | \leq \epsilon$,
we can derive tight bounds relating $Z$ and $\widehat{Z}$.

\begin{align*}
Z &=\int_{I} \exp(S(x))dx
\leq \int_{I} \exp(P(x)+\epsilon)dx \\
&= \hat{Z}\exp(\epsilon) \nn\\	
Z &=\int_{I} \exp(S(x))dx \geq \int_{I} \exp(P(x)-\epsilon)dx \\
&= \hat{Z}\exp(-\epsilon)\nn
\end{align*}

\begin{proof}
\begin{align*}
&D_{KL}(p || \hat{p}) =\int_{I} \ln \left(\frac{p(x)}{\hat{p}(x)} \right)p(x)dx \nn \\		
&= \int_{I}  \left(S(x)-P(x) + \ln(\hat{Z})- \ln(Z) \right)p(x)dx 	\nn	\\
&\leq \int_{I}  \left |S(x)-P(x) \right | p(x)dx \nn \\
& + \int_{I} \left | \ln(\hat{Z})- \ln(Z)  \right |p(x)dx \nn \\
&\leq 2\epsilon \propto \frac{a^{M+1}}{(M+1)!} \approx \frac{1}{\sqrt{2\pi(M+1)!}} \left( \frac{ae}{M+1}  \right)^{M+1}
\end{align*}
where the last approximation follows from Stirling's approximation. Therefore, $D_{KL}(p||\hat{p}) \rightarrow 0$ as $M \rightarrow \infty$. 
\end{proof}

\section{Proof of theorem \ref{thm:exp_family} }
\label{sec:appendix4}
\begin{proof}
\begin{align*}
\log \hat{p}(\theta\mid x_{0:T}) &= \log \left (p(\theta) \prod_{k=0}^T
\hat{p}(x_k | x_{k-1},\theta) \right )\\
&= \log p(\theta) + \sum_{k=0}^T \log \hat{p}(x_k \mid x_{k-1}, \theta) 
\end{align*}
We can calculate the form of $\log \hat{p}(x_k \mid x_{k-1}, \theta)$ explicitly.
\begin{align*}
& \log \hat{p}(x_k \mid x_{k-1}, \theta) = \log \N(\hat{f}(x_{k-1}, \theta), \sigma^2) \\
&= -\log(\sigma \sqrt{2 \pi}) - \frac{(x_k - \hat{f}(x_{k-1}, \theta))^2}{2
\sigma^2} \\
&= -\log(\sigma \sqrt{2 \pi}) - \frac{x_k^2 - 2 x_k \hat{f}(x_{k-1}, \theta) +
\hat{f}(x_{k-1}, \theta)^2}{2 \sigma^2} \\
&= -\log(\sigma \sqrt{2 \pi}) - \frac{x_k^2}{2\sigma^2} - \frac{\sum_{i=0}^M x_k
H^i(x_{k-1}) \theta^i}{\sigma^2} \\
& \qquad + \frac{\sum_{i=0}^{2M} J^i_{x_{k-1}} \theta^i}{2 \sigma^2} 
\end{align*}
Using this expansion, we calculate
{\small
\begin{align*}
\log \hat{p}(\theta\mid x_{0:T}) &=  \log p(\theta) + \sum_{k=0}^T \log
\hat{p}(x_k \mid x_{k-1}, \theta) \\
&= \log p(\theta) - (T+1)\log(\sigma \sqrt{2\pi})\\
&\qquad - \frac{1}{2\sigma^2}\left(\sum_{k=0}^T x_k^2\right) - T(\theta)^T \eta(x_0,\dotsc, x_T) 
\end{align*}
}
where we expand $T(\theta)^T\eta(x_0,\dotsc, x_T)$ as in \ref{thm:exp_family}. The form for $\log \hat{p}(\theta \mid x_{0:T})$ is in the exponential family.
\end{proof}

\section{Proof of theorem \ref{thm:static_error} }
\label{sec:appendix5}
\begin{proof}
Assume that function $f$ has bounded derivatives and bounded support $I$. Then the maximum error
satisfies $\left |
f_{\theta}(x_{k-1})-\hat{f}_{\theta}(x_{k-1}) \right | \leq \epsilon_k $. It
follows that  $\hat{f}_{\theta}(x_{k-1})^2-f_{\theta}(x_{k-1})^2=-\epsilon_k^2-2\hat{f}_{\theta}(x_{k-1})\epsilon_k \approx -2\hat{f}_{\theta}(x_{k-1})\epsilon_k$. 

Then the KL-divergence between the real posterior and the
approximated posterior satisfies the following formula. 
\begin{align}
&D_{KL}(p_T||\hat{p}_T) \\
&\quad= \int_{\mathcal{S}_{\theta}}\left(
\frac{1}{\sigma^2}\sum_{k=1}^T \epsilon_k(x_k-\hat{f}_{\theta}(x_	{k-1}))
\right)p_T(\theta| x_{0:T})d\theta \nn
\end{align}
Moreover, recall that as $T \rightarrow \infty$ the posterior shrinks to
$\delta(\theta-\theta^{*})$ by the assumption of identifiability. Then we can
rewrite the KL-divergence as (assuming Taylor approximation centered around $\theta_c$)
\begin{align}
&\lim_{T \rightarrow \infty}D_{KL}(p_T||\hat{p}_T) \\
&=\frac{1}{\sigma^2} \lim_{T \to \infty} \sum_{k=1}^T \epsilon_k
\int_{\mathcal{S}_{\theta}}(x_k-\hat{f}_{\theta}(x_{k-1}))p_T(\theta| x_{0:T})d\theta \nn \\
&= \frac{1}{\sigma^2} \lim_{T \to \infty} \sum_{k=1}^T \epsilon_k \cdot \\
&\qquad \left( x_k - \sum_{i=0}^M H^i(x_{k-1})\int_{\mathcal{S}_{\theta}}(\theta-\theta_c)^i p(\theta| x_{0:T})d\theta\right) \nn \\
&=\frac{1}{\sigma^2} \lim_{T \to \infty} \sum_{k=1}^T \epsilon_k \left( x_k - \sum_{i=0}^M H^i(x_{k-1})(\theta^{*}-\theta_c)^i \right) \nn
\end{align}

If the center of the Taylor approximation $\theta_c$ is the true parameter value
$\theta^*$, we can show that
\begin{align}
\lim_{T \rightarrow \infty}D_{KL}(p_T||\hat{p}_T) 
&= \frac{1}{\sigma^2}\lim_{T \to \infty} \sum_{k=1}^T \epsilon_k \left( x_k - f_{\theta^*}(x_{k-1})) \right) \nn \\
&= \frac{1}{\sigma^2}\lim_{T \to \infty} \sum_{k=1}^T \epsilon_k v_k =0
\end{align}
where the final statement follows from law of large numbers. Thus, as $T
\rightarrow \infty$, the  Taylor approximation of any order will converge to the
true posterior given that $\theta_c = \theta^*$. For an arbitrary center value $\theta_c$,
{\small
\begin{align}
D_{KL}(p_T||\hat{p}_T) &= \frac{1}{\sigma^2}\sum_{k=1}^T \epsilon_k \left( x_k - \sum_{i=0}^M H^i(x_{k-1}) (\theta^*-\theta_c)^i \right)
\end{align}
}

Notice that $\epsilon_k \propto \frac{1}{(M+1)!}$ (by our assumptions that $f$
has bounded derivative and is supported on interval $I$) and $H^{i}(\cdot) \propto \frac{1}{M!}$. The inner summation will be bounded since $M!>a^M, \forall a \in \mathbb{R}$ as $M \rightarrow \infty$. Therefore, as $M\rightarrow \infty$, $D_{KL}(p||\hat{p})\rightarrow 0$.
\end{proof}

\end{document}